# Trends in AI and Human-AI Interaction in Clinical Trials – A Hybrid Human-AI Exploration

Sandra WOOLLEY [a,1], Tim COLLINS [b], Khalid KHATTAK [a],
Illia CHERNOMORETS [a], Ariane AREVALO [a] and Chris RICHARDSON [a]

[a] *Keele University, UK*
[b] *Manchester Metropolitan University, UK*

ORCiD ID: Sandra Woolley https://orcid.org/0000-0002-7623-2866,
Tim Collins https://orcid.org/0000-0003-2841-1947,
Khalid Khattak https://orcid.org/0000-0001-8410-5915,
Illia Chernomorets https://orcid.org/0009-0004-8871-4012,
Ariane Arevalo https://orcid.org/0009-0002-8068-7529 and
Chris Richardson https://orcid.org/0009-0000-0555-8626

**Abstract.** This paper examines records retrieved from the ClinicalTrials.gov registry to characterize temporal trends in AI terminology and the geographical distribution of AI trials. The work also reports on an exploratory hybrid human-AI approach to analyzing human-AI interaction trends in registered clinical trials. The hybrid workflow comprised a frontier generative AI model (GPT-5.5) and human review to screen and categorize records returned by an AI-focused search. The findings indicate a marked increase in AI-related trials over time, with recent growth in references to machine learning, deep learning, chatbots, GPTs, and large language models. Geographically, China and the United States accounted for the largest numbers of AI-related trials, with notable recent increases in several other countries including Italy, France, Spain, the UK and Turkey (Türkiye). In a random sample of 100 records, human and AI classifiers showed good agreement in identifying studies not substantively using AI, but lower agreement in classifying human-AI interaction, particularly where health professional interaction was ambiguous or insufficiently described. Overall, the results suggest that hybrid human-AI screening of clinical trial records is potentially viable, but clearer trial reporting and more precise interaction definitions will benefit the process.

**Keywords.** Human-AI interaction, clinical trials, hybrid human-AI review

## 1. Introduction

Artificial Intelligence (AI) is increasingly used in clinical research and healthcare practice [1,2]. In clinical trials, AI use can be observed in the publicly available records of trial registries. ClinicalTrials.gov is the largest international registry of clinical trials, documenting trials since 2000 and currently comprising more than 580,000 records for trials located in over 220 countries and territories [3]. AI-related trials may nevertheless be difficult to robustly identify and compare because registry records and published reports vary substantially in how they describe AI methods, define the role of human users, specify input and output data, and characterize the clinical context of deployment [4,5]. SPIRIT-AI (Standard Protocol Items: Recommendations for Interventional Trials-AI) and CONSORT-AI (CONsolidated Standards of Reporting Trials) reporting guidance have sought to improve this by extending standard trial-reporting guidance for AI interventions: SPIRIT-AI recommendations cover protocols and CONSORT-AI recommendations cover completed randomized trial reports [4,5]. However, the existence of reporting guidance has not resolved all of the relevant problems. Evidence from published randomized controlled trials (RCTs) suggests that compliance with CONSORT-AI recommendations remains incomplete, indicating that important features of AI interventions are still underreported in the literature [6].

[1] Corresponding Author: Sandra Woolley, s.i.woolley@keele.ac.uk

Reporting problems are not unique to AI trials. Related work on digital and wearable interventions has similarly identified incomplete specification of intervention technologies, limited version reporting, and broader concerns regarding performance, fairness and transparency within digital health ecosystems [7-9], including concerns about the equitable performance and future deployment of wearable and connected health technologies [10]. This wider context is relevant because AI interventions are often embedded within broader socio-technical systems rather than operating in isolation.

The human-AI interaction dimension is also increasingly significant because the effects of AI in healthcare depend not only on model performance but also on how systems are introduced and integrated into workflows and how users interact with them. Human-AI interaction has been conceptualized in terms of intermittent, continuous and proactive forms of interaction, depending on when and how individuals encounter AI during a task or decision process [11]. In healthcare, however, frameworks for understanding human-AI interaction in real-world settings remain underdeveloped, particularly outside controlled laboratory environments [12]. A further complication is that the meaning of “AI” has evolved over time. Earlier accounts of AI in medicine have emphasized symbolic and rule-based approaches, expert systems and decision-support tools, whereas more recent literature places greater weight on machine learning, deep learning, natural language processing and, more recently still, large language models and conversational systems [2,13].

In addition to its use in healthcare and other domains, AI has been increasingly used to support the time-consuming processes of systematic reviews, from identification and screening prioritization [14,15], with more recent work moving towards human-overseen automation and AI-enhanced review workflows [16,17].

This study examines AI-related clinical trials registered in the ClinicalTrials.gov repository to identify temporal trends, patterns in AI terminology, geographical location, and the extent to which trials can be meaningfully categorized by forms of human-AI interaction. It further explores the feasibility of an experimental AI-assisted screening approach, combined with human checking, for a prospective acceptance-sampling strategy.

## 2. Related Work

Previous studies have identified growth in the registration and reporting of AI-related clinical research, while also observing variations in terminology, application areas and reporting practices.

Maru et al. [18] examined AI and machine learning studies registered on ClinicalTrials.gov between 2010 and 2023 and observed a substantial increase in the number of registered AI-related trials over the period. In a review of RCTs evaluating AI, Han et al. [19] found that published AI trials in clinical practice were concentrated particularly in gastroenterology and radiology, with the United States and China accounting for the largest numbers of trials.

Wekenborg, Gilbert and Kather [12] have argued that the field still lacks sufficiently developed frameworks for assessing human-AI interaction in routine care settings from the user perspective. This is relevant to the study presented in this paper because registry records may not consistently specify how clinicians, patients or other users interact with AI systems.

The literature on conversational AI and large language models adds an important additional consideration. Reviews of medical and healthcare applications of large language models show rapid growth in uses such as medical question answering, dialogue summarization, documentation support, research assistance and clinical decision support [2]. A systematic review of conversational large language models in healthcare likewise reported that these systems are increasingly used for summarization, medical knowledge inquiry, prediction and administrative support, while also raising concerns about reliability, bias, privacy and public acceptability [20].

## 3. Study Aims and Research Questions

The aims of the study were i) to analyze trends in the use of AI and human-AI interaction in clinical trials, and ii) to explore the feasibility of a hybrid human-AI trial screening approach where, prospectively, AI assists the screening process and humans adopted an acceptance-sampling approach. More specifically, the study addressed the following research questions:

*What are the trends in AI clinical trials?*

- What AI terms are used?
- How have these changed over time?
- Where are AI clinical trials located geographically?
- What are the trends in human-AI interaction and conversational-AI chatbots?

*Can trials be categorized in terms of their use of AI and human-AI interaction, i.e., as:*

- No use of AI
- No human-AI interaction
- Patient-AI interaction
- Health professional-AI interaction
- Carer/caregiver-AI interaction
- Hybrid-AI interaction (multi-actor interactions, for example, patients and health professionals both interacting with the AI)
- Other-AI interaction (e.g., medical students)

## 4. Methodology

The ClinicalTrials.gov repository was searched using the AI-focused search string below and applying the inclusion and exclusion criteria. The search string was developed by collating and pruning search terms used in the related literature and by substantial test searches based on names of AI models and methods.

**Search String:**

```
AI OR "artificial intelligence" OR "artificial intelligent" OR chatbot
OR ChatGPT OR GPT OR "computer vision" OR "deep learning" OR "expert
system" OR "large language model" OR LLM OR "machine learning" OR
"natural language processing" OR NLP OR "neural net" OR "neural
network"
```

Each term in the search string was retained on the basis of pilot searches uniquely locating records that other terms did not retrieve. The numbers of records uniquely located by each search term are tabulated in table 1.

**Table 1.** Number of records uniquely located by each search term.

| Search Term | Number of Unique Records |
|---|---|
| AI | 1054 |
| artificial intelligence | 946 |
| machine learning | 784 |
| deep learning | 242 |
| chatbot | 197 |
| GPT | 145 |
| neural network | 145 |
| expert system | 50 |
| computer vision | 49 |
| natural language processing | 19 |
| ChatGPT | 18 |
| large language model | 18 |
| NLP | 16 |
| LLM | 12 |
| artificial intelligent | 8 |
| neural net | 2 |

**Search Fields:** Because the study focused primarily on the *use* of AI within trials an 'Intervention' search was performed. However, observational studies returned by the search were retained and were not excluded on that basis. For ClinicalTrials.gov intervention searches, matches for terms and their synonyms (as defined by the Unified Medical Language System) include the following fields:

- Intervention Name
- Intervention Type
- Arm Group Type
- Intervention Other Name
- Brief Title
- Official Title
- Arm Group Label
- Intervention MeSH Term
- Keyword
- Intervention Ancestor Term
- Intervention Description
- Arm Group Description

**Inclusion Criteria:**
- All clinical trials incorporating identifiable use of AI
- Trials with first submitted dates on or before 1 April 2026

**Exclusion Criteria:**
- No exclusion criteria were applied at the search stage. Records returned by the search were subsequently screened for substantive use of AI.

**Dataset JavaScript Object Notation (JSON) results:** The following fields of included trials from the ClinicalTrials.gov repository were incorporated into the JSON study dataset:
- NCT Number
- Study Title
- Official Title
- Brief Summary
- Detailed Description
- Conditions
- Interventions
- Primary Outcome Measures
- Secondary Outcome Measures
- Other Outcome Measures
- Enrollment
- Study Type
- Study Design

**Classification and Classifiers**: Classification tasks for trials in the JSON dataset required trials to be identified as one of a set of predefined classifications (below) with notes to indicate confidence and uses of chatbots or conversational AI. Interaction was defined as communication or delivery of information during a task or intervention.

- No use of AI
- No human-AI interaction
- Patient AI-interaction
- Caregiver-AI interaction (including paid childminders)
- Health professional-AI interaction
- Other Human-AI interaction (e.g. including medical students)
- Hybrid-AI interaction (more than one group, e.g. patients and health professionals)

The final experimental approach was arrived at iteratively following a series of deep research AI classifications and Python natural language processing (NLP) explorations.
*AI Classification:* A frontier generative AI model (GPT-5.5 accessed via the OpenAI API) was used to perform the screening and categorization. The computation used 10,763,053 input tokens and 3,217,315 output tokens (at a cost of $75.17). The prompt used is provided in the appendix. In addition to classifying trials, a short explanation and a confidence level indication (high, medium or low) were required.
*Human Classification:* Human classifiers, with masters-degree-level expertise or higher in clinical datasets and trial analyses, classified a representative random sample of 100 trials (selected based on keyword results) according to a set of provided classifications. Two human classifiers independently inspected each trial and a third human classifier arbitrated where differences occurred. The two classifiers were allowed to abstain and express uncertainty for trials where they substantially lacked confidence. The resulting human and AI classifications were compared, and agreements and disagreements were then analyzed.

## 5. Results

The AI search string returned 5,828 records for the search conducted on 23 April 2026 for trials first posted on or before 1 April 2026. At the time of writing, this represents slightly over 1% of all ClinicalTrials.gov records.

Of the returned records, 3,019 were interventional studies, 2,807 were observational and two were ‘expanded access’ studies - a special category of trial that enables access to interventions not yet approved for treatment for individuals with serious conditions.

Amongst the sample of 100 AI- and human-classified trials there were nine individual instances where one human classifier abstained from trial classification based on uncertainty. Eight of these were uncertainties regarding human-AI interaction classifications and one regarding the use (or not) of AI.

As summarized in Table 2, human and AI classifiers both identified 14 trials (of the 100 sample trials) as ‘No use of AI’ but differed on the classification of two of these (14%). In these two cases, at least one human classifier abstained or expressed uncertainty, and in both cases the AI classifier reported reduced confidence - the only ‘No use of AI’ AI classifications that were not ‘high’ confidence. Across the whole dataset of trials, the AI classifier identified 958 trials as ‘No use of AI’, 80 of which were labelled as ‘medium’ confidence and 878 as ‘high’ confidence classifications, respectively. Only 20 of the AI classifications were labelled with ‘low’ confidence.

**Table 2.** Categorization counts and agreements between humans and AI classifier.

| Categorization | Human Classifiers | AI Classifier | Agreements |
|---|---|---|---|
| No use of AI | 14 | 14 | 13 |
| No human-AI interaction | 34 | 34 | 22 |
| Patient AI-interaction | 14 | 14 | 10 |
| Caregiver-AI interaction | 2 | 3 | 2 |
| Health professional-AI interaction | 32 | 29 | 20 |
| Other human-AI interaction | 2 | 4 | 2 |
| Hybrid-AI interaction | 2 | 2 | 1 |

Also, as shown in Table 2 human and AI classifier categorizations and agreements for the sample of 100 trials, the classification of human-AI interaction, as distinct from the identification of AI use, proved more challenging. Where there were agreements that AI was used, there were 23 disagreements about whether these involved human-AI interaction. In total, there were 30 classification disagreements, two on the use of AI, 23 on whether interaction was involved, and the remaining five were on the specific category of interaction, e.g. patient-AI interaction vs. health professional-AI interaction. The majority of disagreements regarding the presence of interaction were disagreements specifically between health professional-AI interaction and no human-AI interaction.

As shown in Figure 1, the AI keywords used in trials (AI-classified as using AI) have increased and evolved over the last decade. Though there were some early and notable mentions of AI and “expert system”, there has been a sustained increase in “artificial intelligence” and “AI” since 2016, with similar increases in “machine learning”, “large language model” or “LLM” and “deep learning” since 2017. There has also been a leveling off of some terms including “neural net”/“neural network” and “natural language processing”/“NLP”, while “computer vision” growth appears to continue. Notably, “chatbot” has risen since 2018, and “ChatGPT”/“GPT” has substantially risen since 2022.

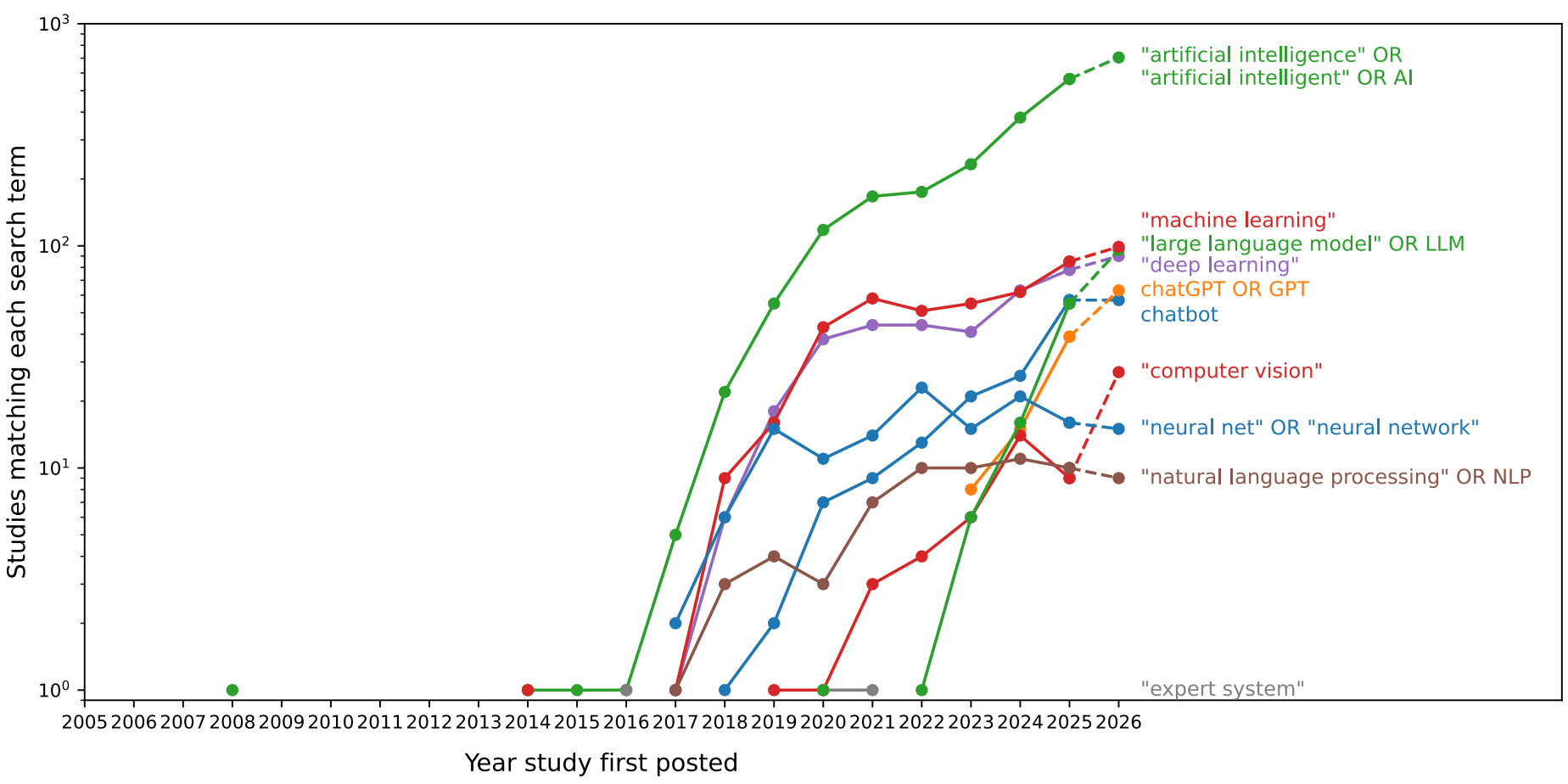

**Figure 1.** Trends in AI terms in AI-classified clinical trials making use of AI.

As shown in Figure 2, the number of trials involving the use of AI has substantially increased in recent years. The number of AI trials located in China increased sharply from 2018, overtaking the number of AI trials located in the USA. Both countries have approximately fourfold as many AI trials as other countries. For example, Italy has the third highest number of AI trials (247) compared to China (989) and the USA (980). Though, as shown, a substantial number of trials did not declare a location. Notable and increasing numbers of AI trials were located in Italy, France, Spain, the UK, Turkey (Türkiye), Taiwan

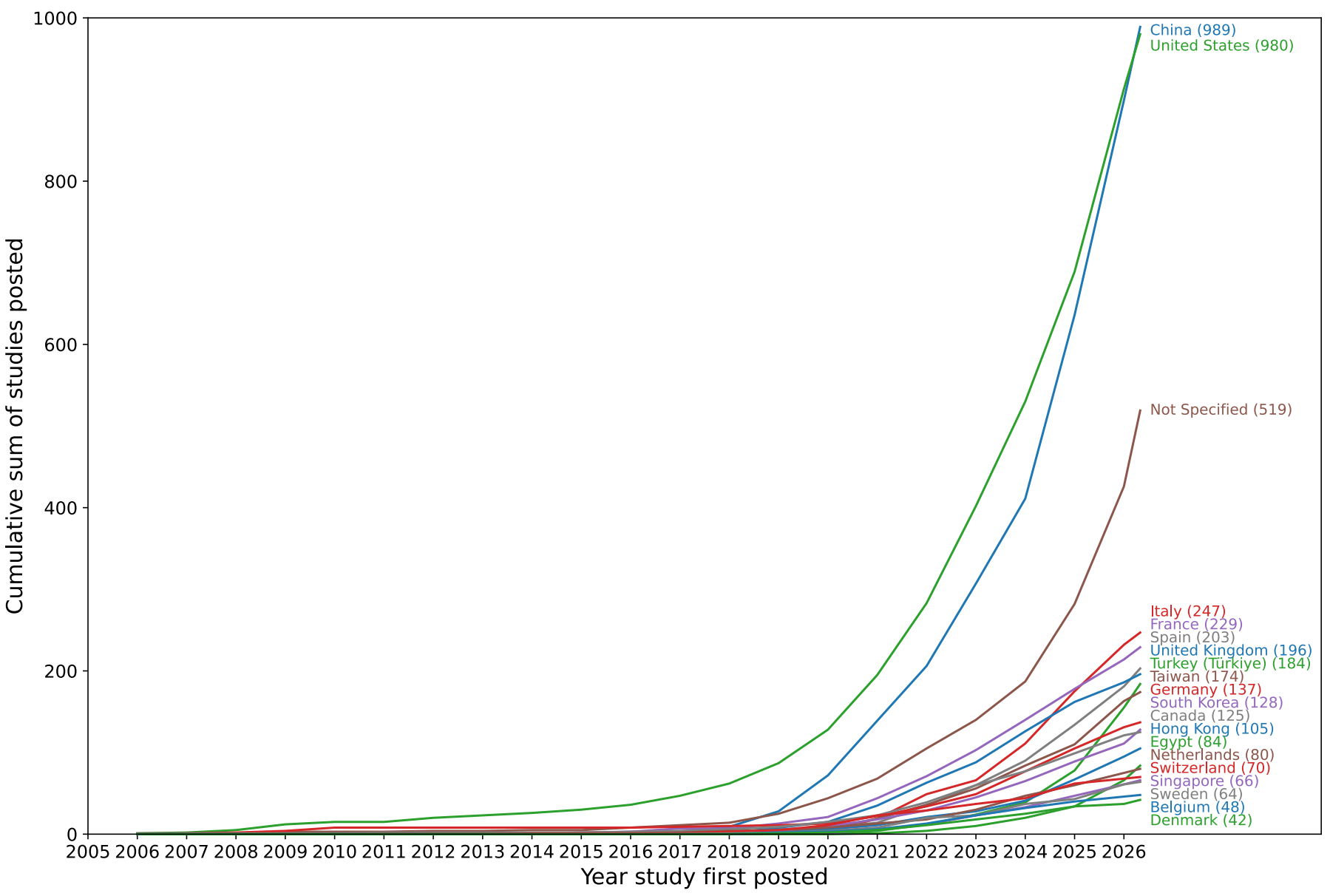

**Figure 2.** Country locations of AI-classified clinical trials making use of AI.

## 6. Discussion

Trials where AI use (vs. no AI use) classification proved challenging included:

i) trials where there was conflicting or ambiguous use of terms across fields, e.g. AI terms such as 'Deep learning technique' included in 'Terms related to this study' but in-documentation references limited to 'automated software' (NCT03206333);

ii) trials where the definition of (substantive use of) AI was insufficiently clear, e.g. where AI was embedded within a component of an intervention, for example, control of sound within a VR experience (NCT04238065).

On inspection of the two less common types of 'expanded access' trials returned by the search string, one made no use of AI and was therefore excluded. The other (NCT05124990) made use of AI but appeared to be an interventional study not aligned with the definition of an expanded access trial.

The computation of the proposed hybrid-human AI approach is not without cost. In addition to the $75.17 API computation fee for the study, pilot explorations largely exhausted the credit of a one-month $200 ChatGPT Pro account. In considering also the environmental cost [21], the total energy consumption for the computation could be equivalent to some tens of kWh and, therefore, one or more kilograms of carbon dioxide ($CO_2$) emissions.

## 7. Conclusions and Further Work

Frontier large language models (such as GPT-5.5) demonstrate potential for screening clinical trials records. A hybrid-human AI approach is, therefore, a potential alternative to the time-intensive human processes involved in systematic reviews of clinical trials and literature, though not without a carbon-footprint.

Attempting to classify modes of human-AI interaction is challenging, particularly when trials are complex and conformance to reporting standards is variable. Also, while there are CONSORT-AI and SPIRIT-AI recommendations to *"Specify whether there was human-AI interaction in the handling of the input data, and what level of expertise was required of users*" [4,5], there are no specific requirements to explicitly categorize human-AI interaction. Future work will include improvements in definitions of interaction, particularly in relation to what does and does not constitute health professional-AI interaction, closer scrutiny of other classification disagreements, and further exploration of programmatic NLP approaches.

In summary, our empirical exploration suggests that there is potential for AI and hybrid human-AI approaches to support the analysis of AI and human-AI interactions in clinical trials, although a more confident assessment will require more detailed and nuanced definitions of interaction and more thorough inspection.

## 8. Acknowledgements

Authors gratefully acknowledge support of the Digital Society Institute of Keele University, UK, that underpins efforts towards the publication of this work. For the purposes of open access, the authors have applied a Creative Commons Attribution (CC-BY) license to any Accepted Author Manuscript version arising from this submission.

## 10. Appendix

AI prompt (markdown format):

```
You are an expert systematic reviewer in clinical applications of AI.

This is a record of a clinical trial taken from the clinicaltrials.gov
repository. Inspect the record and classify the role of artificial
intelligence in the clinical trial using the following categories:

* No use of Artificial Intelligence;
* No human-AI interaction;
* Patient AI-Interaction;
* Caregiver-AI Interaction;
* Health Professional-AI Interaction;
* Other Human-AI Interaction;
* Hybrid-AI Interaction (where interaction involves more than one type
of participants).

Use the trial description, intervention details, population, and other
fields in the record to determine the most appropriate category.

Produce a strict JSON output that has these fields:

* NCT Number
* Artificial Intelligence Interaction Classification Result
* Confidence in the Classification (high, medium, low)
* Reason for Classification
* Concise Summary of AI Use in the Trial
* Chatbot / Conversational AI Identification (true, false)
* Concise summary of Chatbot / Conversational AI Use in the Trial

Complete these fields using concise, research-appropriate wording. In
the 'Chatbot / Conversational AI Identification' field, explicitly
flag whether the trial appears to involve a dialogue with a human and
a chatbot, conversational agent, virtual assistant, dialogue system,
LLM-based interface, or similar technology.

While classifying, follow these principles:

* Do not infer artificial intelligence use from vague digital
terminology alone.
* Distinguish clearly between artificial intelligence used in the
background (i.e. No human-AI interaction) and artificial intelligence
that directly interacts with people. When multiple human groups
interact with the artificial intelligence, use Hybrid-AI Interaction.
* When the evidence is incomplete, keep the explanation cautious and
lower the confidence rating.
* Caregivers can include paid carers such as childminders.
* When AI information is delivered to humans performing a task, this
should be classed as human-AI interaction.
* Medical students should be classed as 'other'.
```